\documentclass[10pt,twocolumn,letterpaper]{article}

\usepackage{cvpr}
\usepackage{times}
\usepackage{epsfig}
\usepackage{graphicx}
\usepackage{amsmath}
\usepackage{amssymb}
\usepackage{multirow}
\usepackage{authblk}
\usepackage{algorithm}

\makeatletter
\renewcommand\AB@affilsepx{ \protect\Affilfont}
\makeatother

\usepackage[pagebackref=true,breaklinks=true,letterpaper=true,colorlinks,bookmarks=false]{hyperref}

\cvprfinalcopy 

\def\cvprPaperID{2448} 

\ifcvprfinal\fi
\begin{document}

\title{CDC: Convolutional-De-Convolutional Networks for Precise Temporal Action Localization in Untrimmed Videos}

\author[$\dagger$]{Zheng Shou}
\author[$\dagger$]{Jonathan Chan}
\author[$\dagger$]{Alireza Zareian}
\author[$\ddagger$]{Kazuyuki Miyazawa}
\author[$\dagger$]{Shih-Fu Chang}
\affil[$\dagger$]{Columbia University, New York, NY, USA;}
\affil[$\ddagger$]{Mitsubishi Electric, Japan}

\affil[ ]{\tt\small \{zs2262,jc4659,az2407,sc250\}@columbia.edu\\
\tt\small Miyazawa.Kazuyuki@cw.mitsubishielectric.co.jp}

\maketitle
\thispagestyle{empty}


\begin{abstract}


Temporal action localization is an important yet challenging problem.
Given a long, untrimmed video consisting of multiple action instances and complex background contents, we need not only to recognize their action categories, but also to localize the start time and end time of each instance.
Many state-of-the-art systems use segment-level classifiers to select and rank proposal segments of pre-determined boundaries.
However, a desirable model should move beyond segment-level and make dense predictions at a fine granularity in time to determine precise temporal boundaries.
To this end, we design a novel \textbf{Convolutional-De-Convolutional} (CDC) network that places CDC filters on top of 3D ConvNets,
which have been shown to be effective for abstracting action semantics but reduce the temporal length of the input data.
The proposed CDC filter performs the required temporal upsampling and spatial downsampling operations simultaneously to predict actions at the frame-level granularity.
It is unique in jointly modeling action semantics in space-time and fine-grained temporal dynamics.
We train the CDC network in an end-to-end manner efficiently.
Our model not only achieves superior performance in detecting actions in every frame, but also significantly boosts the precision of localizing temporal boundaries.
Finally, the CDC network demonstrates a very high efficiency with the ability to process 500 frames per second on a single GPU server.
Source code and trained models are available online at \url{https://bitbucket.org/columbiadvmm/cdc}.


\end{abstract}

\section{Introduction}


Recently, temporal action localization has drawn considerable interest in the computer vision community \cite{THUMOS14,THUMOS15,th1,th2,th3,AN1,AN2,scnn_shou_wang_chang_cvpr16,Richard_2016_CVPR,stanford_cvpr16,victor_eccv16,fast_temporal_activity_cvpr16,spoton_eccv16}.
This task involves two components: (1) determining whether a video contains specific actions (such as diving, jump, \etc) and (2) identifying temporal boundaries (start time and end time) of each action instance.


\begin{figure}[b]
\centering
\includegraphics[width=0.5\textwidth]{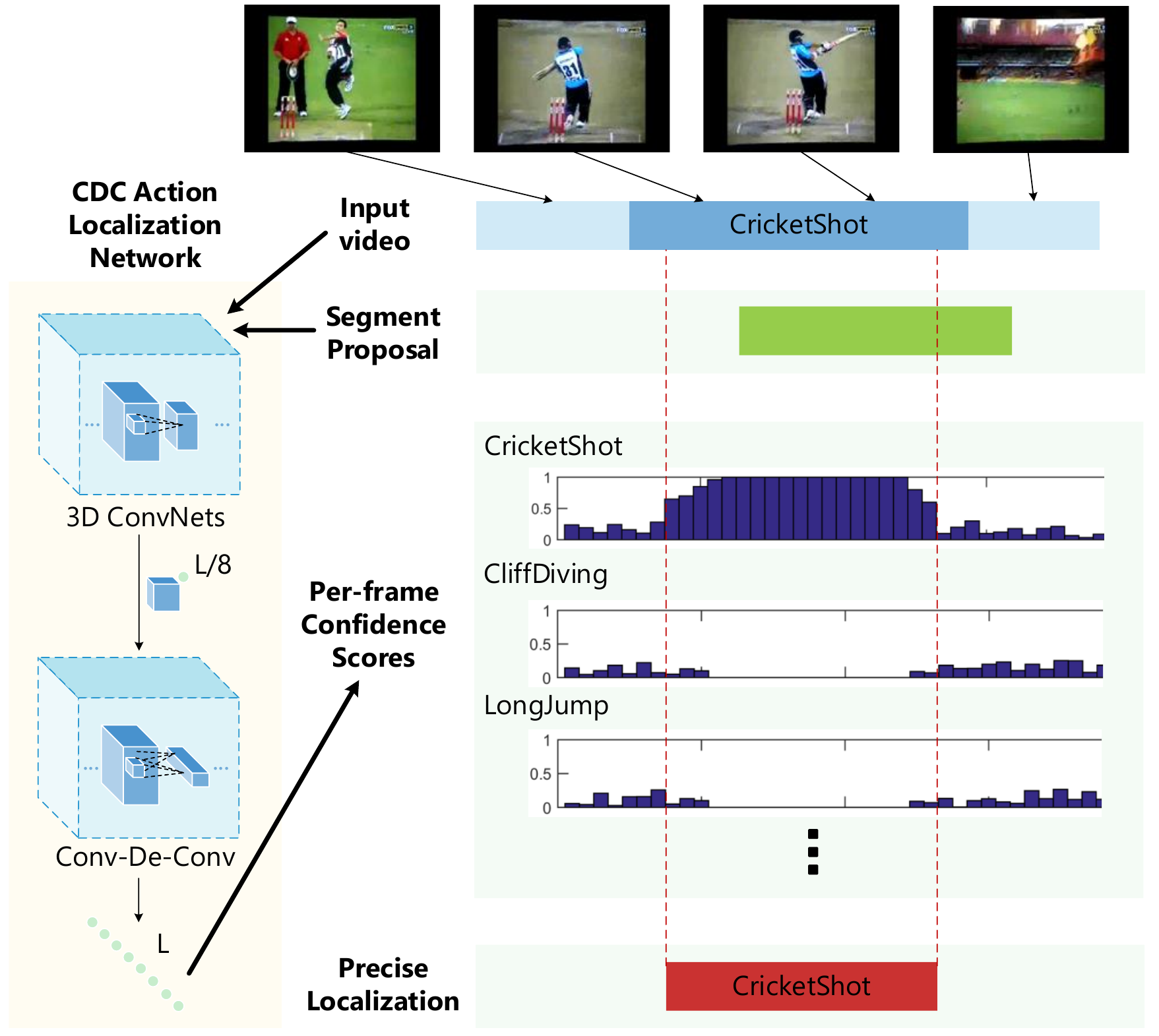}
\caption{
Our framework for precise temporal action localization.
Given an input raw video,
it is fed into our CDC localization network, which consists of 3D ConvNets for semantic abstraction and a novel CDC network for dense score prediction at the frame-level. Such fine-granular score sequences are combined with segment proposals to detect action instances with precise boundaries.
}
\label{framework}
\end{figure}

A typical framework used by many state-of-the-art systems \cite{AN1,AN2,th1,th2,th3} is fusing a large set of features and training classifiers that operate on sliding windows or segment proposals. 
Recently, an end-to-end deep learning framework called Segment-CNN (S-CNN) \cite{scnn_shou_wang_chang_cvpr16} based on 3D ConvNets \cite{3dcnn} demonstrated superior performances both in efficiency and accuracy on standard benchmarks such as THUMOS'14 \cite{THUMOS14}.
S-CNN consists of a proposal network for generating candidate video segments and a localization network for predicting segment-level scores of action classes.
Although the localization network can be optimized to select segments with high overlaps with ground truth action instances,
the detected action boundaries are still retained and thus are restricted to the pre-determined boundaries of a fixed set of proposal segments.



As illustrated in Figure \ref{framework}, our goal is to refine temporal boundaries from proposal segments to precisely localize boundaries of action instances.
This motivates us to move beyond existing practices based on segment-level predictions, and explicitly focus on the issue of fine-grained, dense predictions in time.
To achieve this goal, some existing techniques can be adapted: (1) Single-frame classifiers operate on each frame individually; (2) Recurrent Neural Networks (RNN) further take into account temporal dependencies across frames.
But both of them fail to explicitly model the spatio-temporal information in raw videos.

3D CNN \cite{3dcnn,scnn_shou_wang_chang_cvpr16} has been shown that it can learn spatio-temporal abstraction of high-level semantics directly from raw videos but loses granularity in time, which is important for precise localization, as mentioned above.
For example, layers from $\tt conv1a$ to $\tt conv5b$ in the well-known C3D architecture \cite{3dcnn} reduce the temporal length of an input video by a factor of 8.
In pixel-level semantic segmentation, de-convolution proves to be an effective upsampling method in both image \cite{Long_2015_CVPR,Long_2016_PAMI} and video \cite{V2V} for producing output of the same resolution as the input.
In our temporal localization problem, the temporal length of the output should be the same as the input video, but the spatial size should be reduced to 1x1.
Therefore, we not only need to upsample in time but also need to downsample in space.
To this end, we propose a novel \textbf{Convolutional-De-Convolutional} (CDC) filter, which performs convolution in space (for semantic abstraction) and de-convolution in time (for frame-level resolution) simultaneously.
It is unique in jointly modeling the spatio-temporal interactions between summarizing high-level semantics in space and inferring fine-grained action dynamics in time.
On top of 3D ConvNets, we stack multiple CDC layers to form our CDC network, which can achieve the aforementioned goal of temporal upsampling and spatial downsampling,
and thereby can determine action categories and can refine boundaries of proposal segments to precisely localize action instances.



In summary, this paper makes three novel contributions:

(1) To the best of our knowledge, this is the first work to combine two reverse operations (\ie convolution and de-convolution) into a joint CDC filter, which simultaneously conducts downsampling in space and upsampling in time to infer both high-level action semantics and temporal dynamics at a fine granularity in time.

(2) We build a CDC network using the proposed CDC filter to specifically address precise temporal action localization. The CDC network can be efficiently trained end-to-end from raw videos to produce dense scores that are used to predict action instances with precise boundaries.

(3) Our model outperforms state-of-the-art methods in video per-frame action labeling and significantly boosts the precision of temporal action localization over a wide range of detection thresholds.


\section{Related work}

\noindent\textbf{Action recognition and detection.}
Early works mainly focus on simple actions in well-controlled environments and can be found in recent surveys \cite{survey1,survey2,survey3}. Recently, researchers have started investigating untrimmed videos in the wild and have designed various features and techniques. We briefly review the following that are also useful in temporal action localization: frame-level Convolutional Neural Networks (CNN) trained on ImageNet \cite{ILSVRC15} such as AlexNet  \cite{alex}, VGG \cite{Simonyan15}, ResNet \cite{He_2016_CVPR},  \etc; 3D CNN architecture called C3D \cite{3dcnn} trained on a large-scale sports video dataset \cite{sports1m} ; improved Dense Trajectory Feature (iDTF) \cite{dtf,idtf} consisting of HOG, HOF, MBH features extracted along dense trajectories with camera motion influences eliminated; key frame selection \cite{Gan_2015_CVPR}; ConvNets adapted for using motion flow as input \cite{Simonyan14b,Feichtenhofer_2016_CVPR,TSN};  feature encoding with Fisher Vector (FV) \cite{FV1,Oneata2} and VLAD \cite{VLAD1,xu2015discriminative}. 

There are also studies on spatio-temporal action detection, which aim to detect action regions with bounding boxes over consecutive frames. Various methods have been developed, from the perspective of supervoxel merging \cite{tube,walk,Soomro_2016_CVPR}, tracking \cite{learntrack,tubedt,apt,Singh_2016_CVPR}, object detection and linking \cite{humanfocus,actiontubes,gangyu,tubedt,apt}, spatio-temporal segmentation \cite{ST-CNN,Xu_2016_CVPR}, and leveraging still images \cite{objectaction,Sultani_2016_CVPR,15000}.

\hfill\break\noindent\textbf{Temporal action localization.} Gaidon \etal \cite{actoms2,actoms} introduced the problem of temporally localizing actions in untrimmed videos, focusing on limited actions such as ``drinking and smoking'' \cite{Laptev_2007} and ``open door and sit down'' \cite{Laptev_2009}. Later, researchers worked on building large-scale datasets consisting of complex action categories, such as THUMOS \cite{THUMOS14,THUMOS15} and MEXaction2 \cite{mex1,mex2,mex3}, and datasets focusing on fine-grained actions \cite{MPII,Charades1,Charades2} or activities of high-level semantics \cite{caba2015activitynet}. The typical approach used in most systems \cite{AN1,AN2,th1,th2,th3} is extracting a pool of features, which are fed to train SVM classifiers, and then applying these classifiers on sliding windows or segment proposals for prediction.
In order to design a model specific to temporal localization, Richard and Gall \cite{Richard_2016_CVPR} proposed using statistical length and language modeling to represent temporal and contextual structures. Heilbron \etal \cite{fast_temporal_activity_cvpr16} introduced a sparse learning framework for generating segment proposals of high recall.

Recently, deep learning methods showed improved performance in localizing action instances. 
RNN has been widely used to model temporal state transitions over frames: Escorcia \etal \cite{victor_eccv16} built a temporal action proposal system based on Long-Short Term Memory (LSTM);
Yeung \etal \cite{stanford_cvpr16} used REINFORCE to learn decision policies for a RNN-based agent;
Yeung \etal \cite{yeung2015every} introduced MultiTHUMOS dataset of multi-label annotations for every frame in THUMOS videos and defined a LSTM network to model multiple input and output connections;
Yuan \etal \cite{yuan_cvpr16} proposed a pyramid of score distribution feature at the center of each sliding window to capture the motion information over multiple resolutions, and utilized RNN to improve inter-frame consistency;
Sun \etal \cite{sssn_mm15} leveraged web images to train LSTM model when only video-level annotations are available.
In addition, Lea \etal \cite{ST-CNN} used temporal 1D convolution to capture scene changes when actions were being performed.
Although RNN and temporal 1D convolution can model temporal dependencies among frames and make frame-level predictions, they are usually placed on top of deep ConvNets, which take a single frame as input, rather than directly modeling spatio-temporal characteristics in raw videos.
Shou \etal \cite{scnn_shou_wang_chang_cvpr16} proposed an end-to-end Segment-based 3D CNN framework (S-CNN), which outperformed other RNN-based methods by capturing spatio-temporal information simultaneously.
However, S-CNN lacks the capability to predict at a fine time resolution and to localize precise temporal boundaries of action instances.
\label{soa}

\hfill\break\noindent\textbf{De-convolution and semantic segmentation.} Zeiler \etal \cite{deconv_cvpr10} originally proposed de-convolutional networks for image decomposition, and later Zeiler and Fergus \cite{deconv_eccv14} re-purposed de-convolutional filter to map CNN activations back to the input to visualize where the activations come from.
Long \etal  \cite{Long_2015_CVPR,Long_2016_PAMI} showed that deep learning based approaches can significantly boost performance in image semantic segmentation.
They proposed Fully Convolutional Networks (FCN) to output feature maps of reduced dimensions, and then employed de-convolution for upsampling to make dense, pixel-level predictions.
The fully convolutional architecture and learnable upsampling method are efficient and effective, and thus inspired many extensions \cite{Noh_2015_ICCV,hong2015decoupled, Liu_2016_CVPR,SegNet_2016_PAMI,Lin_2016_CVPR,Zheng_2015_ICCV,deeplab_2015,deeplab_2016,YuKoltun2016}. 

Recently, Tran \etal \cite{V2V} extended de-convolution from 2D to 3D and achieved competitive results on various voxel-level prediction tasks such as video semantic segmentation. This shows that de-convolution is also effective in the video domain and has the potential to be adapted for making dense predictions in time for our temporal action localization task.
However, unlike the problem of semantic segmentation, we need to upsample in time but maintain downsampling in space. Instead of stacking a convolutional layer and a de-convolutional layer to conduct upsampling and downsampling separately, our proposed CDC filter learns a joint model to perform these two operations simultaneously, and proves to be more powerful and easier to train.


\section{Convolutional-De-Convolutional networks}\label{cdc}

\subsection{The need of downsampling and upsampling} \label{build}

C3D architecture, consisting of 3D ConvNets followed by three Fully Connected (FC) layers, has achieved promising results in video analysis tasks such as recognition \cite{3dcnn} and localization \cite{scnn_shou_wang_chang_cvpr16}.
Further, Tran \etal \cite{V2V} experimentally demonstrated the 3D ConvNets, \ie from $\tt conv1a $ to $\tt conv5b $, to be effective in summarizing spatio-temporal patterns from raw videos into high-level semantics.

Therefore, we build our CDC network upon C3D.
We adopt from $\tt conv1a $ to $\tt conv5b $ as the first part of our CDC network.
For the rest of layers in C3D,
we keep $\tt pool5$ to perform max pooling in height and width by a factor of 2 but retain the temporal length.
Following conventional settings \cite{3dcnn,scnn_shou_wang_chang_cvpr16,V2V}, we set the height and width of the CDC network input to 112x112.
Given an input video segment of temporal length $L$, the output data shape of $\tt pool5 $ is (512, $L/8$, 4, 4) \protect\footnotemark \footnotetext{We denote \label{notate} the shape of data in the networks using the form of (number of channels, temporal length, height, width) and the size of feature map, kernel, stride, zero padding using (temporal length, height, width).}.
Now in order to predict the action class scores at the original temporal resolution (frame-level), we need to upsample in time (from $L/8$ back to $L$), and downsample in space (from 4x4 to 1x1).
To this end, we propose the CDC filter and design a CDC network to adapt the FC layers from C3D to perform the required upsample and downsample operations. Details are described in Sections \ref{3.2} and \ref{3.3}.

\subsection{CDC filter} \label{3.2}

In this section, we walk through a concrete example of adapting $\tt FC6$ layer in C3D to perform spatial downsampling by a factor of 4x4 and temporal upsampling by a factor of 2.
For the sake of clarity, we focus on how a filter operates within one input channel and one output channel.

\begin{figure}[b]
\centering
\includegraphics[width=0.46\textwidth]{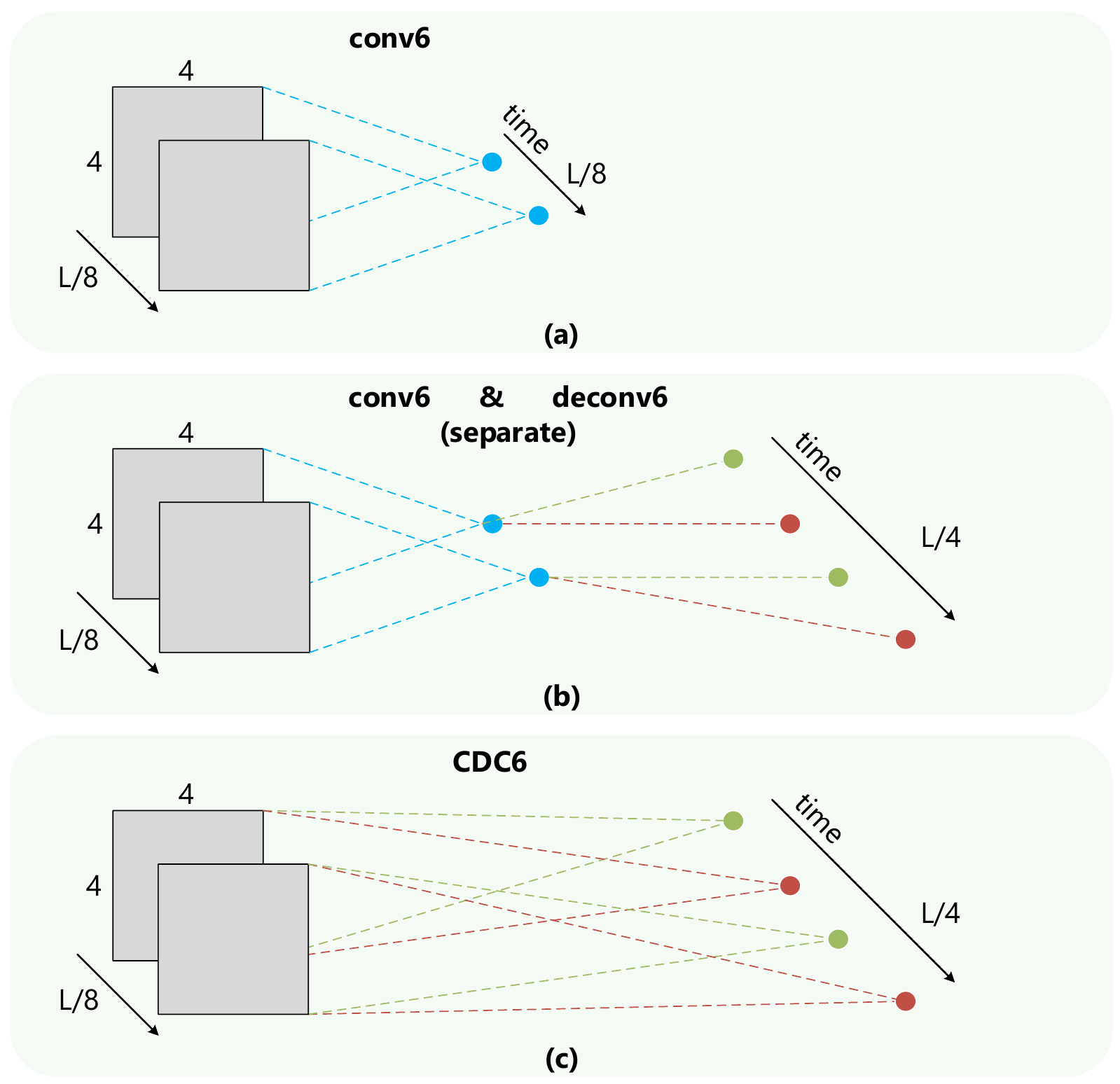}
\caption{Illustration of how a filter in $\tt conv6$, $\tt deconv6$, $\tt CDC6$ operates on $\tt pool5$ output feature maps (grey rectangles) stacked in time. In each panel, dashed lines with the same color indicate the same filter sliding over time. Nodes stand for outputs.}
\label{filter}
\end{figure}

As explained in \cite{Long_2015_CVPR,Long_2016_PAMI}, the FC layer is a special case of a convolutional layer (when the input data and the kernel have the same size and there is no striding and no padding). 
So we can transform $\tt FC6$ into $\tt conv6$, which is shown in Figure \ref{filter} (a).
Previously, a filter in $\tt FC6$ takes a 4x4 feature map from $\tt pool5 $ as input and outputs a single value.
Now, a filter in $\tt conv6$ can slide on $L/8$ feature maps of size 4x4 stacked in time and respectively output $L/8$ values in time. The kernel size of $\tt conv6$ is 4x4=16.

Although $\tt conv6$ performs spatial downsampling, the temporal length remains unchanged. To upsample in time, as shown in Figure \ref{filter} (b), a straightfoward solution adds a de-convolutional layer $\tt deconv6$ after $\tt conv6$ to double the temporal length while maintaining the spatial size. The kernel size of $\tt deconv6$ is 2. Therefore, the total number of parameters for this solution (separated $\tt conv6$ and $\tt deconv6$) is 4x4+2=18.

However, this solution conducts temporal upsampling and spatial downsampling in a separate manner.
Instead, we propose the CDC filter $\tt CDC6$ to jointly perform these two operations. As illustrated in Figure \ref{filter} (c), a $\tt CDC6$ filter consists of two independent convolutional filters (the red one and the green one) operating on the same input 4x4 feature map.
Each of these convolutional filters has the same kernel size as the filter in $\tt conv6$ and separately outputs one single value. So each 4x4 feature map results in 2 outputs in time.
As the CDC filter slides on $L/8$ feature maps of size 4x4 stacked in time, this input feature volume of temporal length $L/8$ is upsampled in time to $L/4$, and its spatial size is reduced to 1x1.
Consequently, in space this CDC filter is equivalent to a 2D convolutional filter of kernel size 4x4; in time it has the same effect as a 1D de-convolutional filter of kernel size 2, stride 2, padding 0.
The kernel size of such a joint filter in $\tt CDC6$ is 2x4x4=32, which is larger than the separate convolution and de-convolution solution (18).

Therefore, a CDC filter is more powerful for jointly modeling high-level semantics and temporal dynamics: each output in time comes from an independent convolutional kernel dedicated to this output (the red/green node corresponds to the red/green kernel); however, in the separate convolution and de-convolution solution, different outputs in time share the same high-level semantics (the blue node) outputted by one single convolutional kernel (the blue one).

Having more parameters makes the CDC filter harder to learn. To remedy this issue, we propose a method to adapt the pre-trained $\tt FC6$ layer in C3D to initialize $\tt CDC6$. After we convert $\tt FC6$ to $\tt conv6$, $\tt conv6$ and $\tt CDC6$ have the same number of channels (\ie 4,096) and thus the same number of filters.
Each filter in $\tt conv6$ can be used to initialize its corresponding filter in $\tt CDC6$:
the filter in $\tt conv6$ (the blue one) has the same kernel size as each of these two convolutional filters (the red one and the green one) in the $\tt CDC6$ filter and thus can serve as the initialization for them both.\label{init}

Generally, assume that a CDC filter $F$ of kernel size ($k_l$, $k_h$, $k_w$) takes the input receptive field $X$ of height $k_h$ and width $k_w$, and produces $Y$ that consists of $k_l$ successive outputs in time.
For the example given in Figure \ref{filter} (c), we have $k_l=2$, $k_h=4$, $k_w=4$.
Given the indices $a \in \left\{ {1,...,{k_h}} \right\} $ and $b \in \left\{ {1,...,{k_w}} \right\}$ in height and width respectively for $X$ and the index $c \in \left\{ {1,...,{k_l}} \right\}$ in time for $Y$: during the forward pass, we can compute $Y$ by \begin{equation} Y\left[ c \right] = \sum\limits_{a = 1}^{{k_h}} {\sum\limits_{b = 1}^{{k_w}} {F\left[ {c,a,b} \right] \cdot X\left[ {a,b} \right]} }; \end{equation} during the back-propagation, our CDC filter follows the chain rule and propagates gradients from $Y$ to $X$ via \begin{equation} X\left[ {a,b} \right] = \sum\limits_{c = 1}^{{k_l}} {F\left[ {c,a,b} \right] \cdot } Y\left[ c \right]. \end{equation}
A CDC filter $F$ can be regarded as coupling a series of convolutional filters (each one has kernel size $k_h$ in height and $k_w$ in width) in time with a shared input receptive field $X$, and at the same time, $F$ performs 1D de-convolution with kernel size $k_l$ in time.
In addition, the cross-channel mechanisms within a CDC layer and the way of adding biases to the outputs of the CDC filters follow the conventional strategies used in convolutional and de-convolutional layers.

\begin{figure*}[t]
\centering
\includegraphics[width=\textwidth]{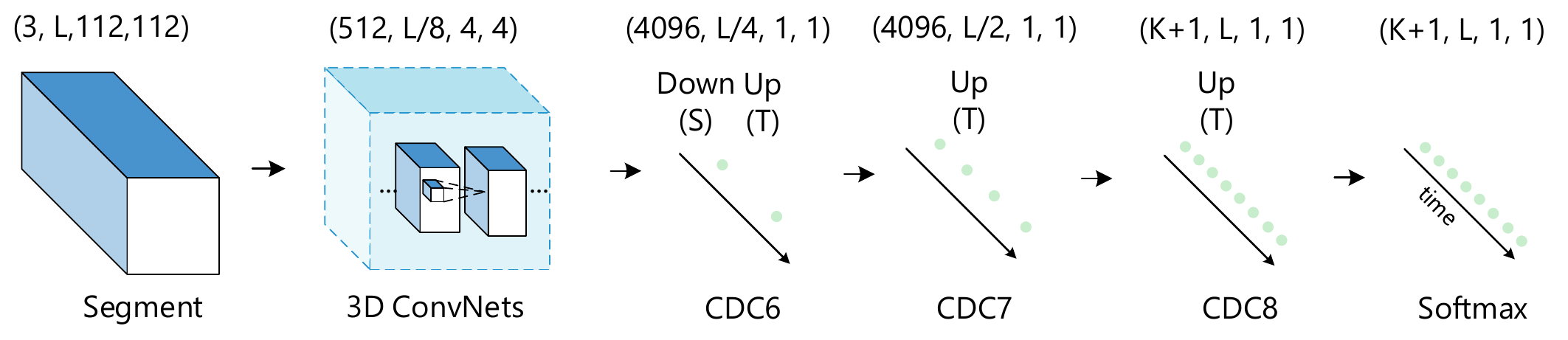}
\caption{Architecture of a typical CDC network. Following the notations indicated in the footnote \ref{notate}, the top row lists the shape of output data at each layer. (1) A video segment is first fed into 3D ConvNets and the temporal length reduces from $L$ to $L/8$.
(2) $\tt CDC6 $ has kernel size (4, 4, 4), stride (2, 1, 1), padding (1, 0, 0), and therefore reduces both height and width to 1 while increases the temporal length from $L/8$ to $L/4$. Both $\tt CDC7 $ and $\tt CDC8 $ have kernel size (4, 1, 1), stride (2, 1, 1), padding (1, 0, 0), and hence both $\tt CDC7 $ and $\tt CDC8 $ further perform upsampling in time by a factor of 2, and thus the temporal length is back to $L$.
(3) A frame-wise softmax layer is added on top of $\tt CDC8 $ to obtain confidence scores for every frame. Each channel stands for one class.
}
\label{network}
\end{figure*}

\subsection{Design of CDC network architecture} \label{3.3}

In Figure \ref{network}, we illustrate our CDC network for labeling every frame of a video.
The final output shape of the CDC network is ($K$+1, $L$, 1, 1), where $K$+1 stands for $K$ action categories plus the background class.
As described  in Section \ref{build}, from $\tt conv1a $ to $\tt pool5 $, the temporal length of an input segment has been reduced from $L$ to $L/8$.
On top of $\tt pool5 $, in order to make per-frame predictions, we adapt FC layers in C3D as CDC layers to perform temporal upsampling and spatial downsampling operations.
Following previous de-convolution works \cite{V2V,Long_2015_CVPR,Long_2016_PAMI}, we upsample in time by a factor of 2 in each CDC layer, to gradually increase temporal length from $L/8$ back to $L$.

In the previous Section \ref{init}, we provide an example of how to adapt $\tt FC6$ as $\tt CDC6 $,
performing temporal 1D de-convolution of kernel size 2, stride 2, padding 0.
For $\tt CDC6 $ in the CDC network, we construct a CDC filter with 4 convolutional filters instead of 2, and thus its temporal kernel size in time increases from 2 to 4. We set the corresponding stride to 2 and padding to 1.
Now each 4x4 feature map produces 4 output nodes, and every two consecutive feature maps have 2 nodes overlapping in time.
Consequently, the temporal length of input is still upsampled by $\tt CDC6 $ from $L/8$ to $L/4$, but each output node sums contributions from two consecutive input feature maps, allowing temporal dynamics in input to be taken into account.

Likewise, we can adapt $\tt FC7$ as $\tt CDC7$, as indicated in Figure \ref{network}.
Additionally, we retain the Relu layers and the Dropout layers with 0.5 dropout ratio from C3D to attach to both $\tt CDC6 $ and  $\tt CDC7 $.
$\tt CDC8 $ corresponds to $\tt FC8$ but cannot be directly adapted from $\tt FC8$ because the classes in $\tt FC8$ and $\tt CDC8 $ are different.
Since each channel stands for one class, $\tt CDC8 $ has $K$+1 channels.
Finally, the $\tt CDC8 $ output is fed into a frame-wise softmax layer $\tt  Softmax$ to produce per-frame scores.
During each mini-batch with $N$ training segments, for the $n$-th segment, the $\tt CDC8 $ output $O_n$ has the shape ($K$+1, $L$, 1 ,1).
For each frame, performing the conventional softmax operation and computing the softmax loss and gradient are independent of other frames.
Corresponding to the $t$-th frame, the $\tt CDC8 $ output ${O_n}\left[ t \right]$ and $\tt  Softmax$ output ${P_n}\left[ t \right]$ both are vectors of $K$+1 values.
Note that for the $i$-th class, $P_n^{\left( i \right)}\left[ t \right] = \frac{{{e^{O_n^{\left( i \right)}\left[ t \right]}}}}{{\sum\nolimits_{j = 1}^{K+1} {{e^{O_n^{\left( j \right)}\left[ t \right]}}} }}$.
The total loss ${\cal L}$ is defined as:
\begin{equation} {\cal L} = \frac{1}{N}\sum\limits_{n = 1}^N {\sum\limits_{t = 1}^L {\left( { - \log \left( {P_n^{\left( {{z_n}} \right)}\left[ t \right]} \right)} \right)} }, \end{equation} where $z_n$ stands for the ground truth class label for the $n$-th segment.
The total gradient w.r.t the output of $i$-th channel/class and $t$-th frame in $\tt CDC8$ is the summation over all $N$ training segments of:
\begin{equation}
\frac{{\partial {\cal L}}}{{\partial O_n^{\left( i \right)}\left[ t \right]}} = \left\{ {\begin{array}{*{20}{c}}
{\frac{1}{N} \cdot \left( {P_n^{\left( {{z_n}} \right)}\left[ t \right] - 1} \right)}&{{\rm{if}}\:i = {z_n}}\\
{\frac{1}{N} \cdot P_n^{\left( i \right)}\left[ t \right]}&{{\rm{if}}\:i \ne {z_n}}
\end{array}} \right.
.\end{equation}

\subsection{Training and prediction}


\noindent\textbf{Training data construction.} In theory, because both the convolutional filter and the CDC filter slide over the input, they can be applied to input of arbitrary size. Therefore, our CDC network can operate on videos of variable lengths.
Due to GPU memory limitations, in practice we slide a temporal window of 32 frames without overlap on the video and feed each window individually into the CDC network to obtain dense predictions in time.
From the temporal boundary annotations, we know the label of every frame.
Frames in the same window can have different labels.
To prevent including too many background frames for training, we only keep windows that have at least one frame belonging to actions.
Therefore, given a set of training videos, we obtain a training collection of windows with frame-level labels.


\hfill\break\noindent\textbf{Optimization.} We use stochastic gradient descent to train the CDC network with the aforementioned frame-wise softmax loss. Our implementation is based on Caffe \cite{caffe} and C3D \cite{3dcnn}. The learning rate is set to 0.00001 for all layers except for $\tt CDC8 $ layer where the learning rate is 0.0001 since $\tt CDC8 $ is randomly initialized.
Following conventional settings \cite{3dcnn,scnn_shou_wang_chang_cvpr16}, we set momentum to 0.9 and weight decay to 0.005.

C3D \cite{3dcnn} is trained on Sports-1M \cite{sports1m} and can be used to directly initialize $\tt conv1a $ to $\tt conv5b $. $\tt CDC6 $ and  $\tt CDC7 $ are initialized by $\tt FC6 $ and  $\tt FC7 $ respectively using the strategy described in the Section \ref{init}. In addition, since $\tt FC8 $ in C3D and $\tt CDC8 $ in the CDC network have the different number of channels, we randomly initialize $\tt CDC8 $. With such initialization, our CDC network turns out to be very easy to train and converges quickly, \ie 4 training epochs (within half a day) on THUMOS'14 .


\hfill\break\noindent\textbf{Fine-grained prediction and precise localization.} \label{post} During testing, after applying the CDC network on the whole video, we can make predictions for every frame of the video.
Through thresholding on confidence scores and grouping adjacent frames of the same label, it is possible to cut the video into segments and produce localization results. But this method is not robust to noise, and designing temporal smoothing strategies turns out to be ad hoc and non-trivial.
Recently, researchers developed some efficient segment proposal methods \cite{scnn_shou_wang_chang_cvpr16,victor_eccv16} to generate a small set of candidate segments of high recall.
Utilizing these proposals for our localization model not only bypasses the challenge of grouping adjacent frames, but also achieves considerable speedup during testing, because we only need to apply the CDC network on the proposal segments instead of the whole video.

Since these proposal segments only have coarse boundaries, we propose using fine-grained predictions from the CDC network to localize precise boundaries.
First, to look at a wider interval, we extend each proposal segment's boundaries on both sides by the percentage $\alpha$ of the original segment length.
We set $\alpha$ to 1/8 for all experiments.
Then, similar to preparing training segments, we slide temporal windows without overlap on the test videos.
We only need to keep test windows that overlap with at least one extended proposal segment.
We feed these windows into our CDC network and generate per-frame action classes scores.

The category of each proposal segment is set to the class with the maximum average confidence score over all frames in the segment. If a proposal segment does not belong to the background class, we keep it and further refine its boundaries.
Given the score sequence of the predicted class in the segment, we perform Gaussian kernel density estimation and obtain its mean $\mu$ and standard deviation $\sigma$.
Starting from the boundary frame at each side of the extended segment and moving towards its middle, we shrink its temporal boundaries until we reach a frame with the confidence score no lower than $\mu$ - $\sigma$. 
Finally, we set the prediction score of the segment to the average confidence score of the predicted class over frames in the refined segment of  boundaries.

\begin{table}[b]
\begin{center}
\begin{tabular}{c|c}
methods & mAP \\ \hline
Single-frame CNN \cite{Simonyan15} & 34.7  \\
Two-stream CNN \cite{Simonyan14b} & 36.2  \\
LSTM  \cite{lrcn2014} & 39.3  \\
MultiLSTM  \cite{yeung2015every} & 41.3  \\ \hline
C3D + LinearInterp &  37.0  \\
Conv \& De-conv &  41.7 \\
CDC (fix 3D ConvNets) &  37.4  \\   \hline
\textbf{CDC}   &    \textbf{44.4}
\end{tabular}
\end{center}
\caption{Per-frame labeling mAP on THUMOS'14 .}
\label{map1}
\end{table}

\section{Experiments} \label{exp}


\subsection{Per-frame labeling}

We first demonstrate the effectiveness of our model in predicting accurate labels for every frame.
Note that this task can accept an input of multiple frames to take into account temporal information.
We denote our model as \textbf{CDC}.

\hfill\break\noindent\textbf{THUMOS'14 \cite{THUMOS14}.} The temporal action localization task in THUMOS Challenge 2014 involves 20 actions. We use 2,755 trimmed training videos and 1,010 untrimmed validation videos (3,007 action instances) to train our model. For testing, we use all 213 test videos (3,358 action instances) which are not entirely background videos.

\hfill\break\noindent\textbf{Evaluation metrics.} \label{eval1}
Following conventional metrics \cite{yeung2015every}, we treat the per-frame labeling task as a retrieval problem. For each action class, we rank all frames in the test set by their confidence scores for that class and compute Average Precision (AP). Then we average over all classes to obtain mean AP (mAP).

\hfill\break\noindent\textbf{Comparisons.} In Table \ref{map1}, we first compare our CDC network (denoted by CDC) with some state-of-the-art models (results are quoted from \cite{yeung2015every}):
(1) Single-frame CNN: the frame-level 16-layer VGG CNN model \cite{Simonyan15};
(2) Two-stream CNN: the frame-level two-stream CNN model proposed in \cite{Simonyan14b}, which has one stream for pixel and one stream for optical flow;
(3) LSTM: the basic per-frame labeling LSTM model of 512 hidden units \cite{lrcn2014} on the top of VGG CNN $ \tt FC7$ layer;
(4) MultiLSTM: a LSTM model developed by Yeung \etal \cite{yeung2015every} to process multiple input frames together with temporal attention mechanism and output predictions for multiple frames.
Single-frame CNN only takes into account appearance information.
Two-stream CNN models appearance and motion information separately. LSTM based models can capture temporal dependencies across frames but do not model motion explicitly. Our \textbf{CDC} model is based on 3D convolutional layers and CDC layers, which can operate on spatial and temporal dimensions simultaneously, achieving the best performance.

In addition, we compare CDC with other C3D based approaches that use different upsampling methods.
(1) C3D + LinearInterp: we train a segment-level C3D using the same set of training segments whose segment-level labels are determined by the majority vote. During testing we perform linear interpolation to upsample segment-level predictions as frame-level.
(2) Conv \& De-conv: $\tt CDC7$ and $\tt CDC8$ in our CDC network keep the spatial data shape unchanged and therefore can be also regarded as de-convolutional layers. For $\tt CDC6$, we replace it with a convolutional layer $\tt conv6$ and a separate de-convolutional layer $\tt deconv6$ as shown in Figure \ref{filter} (b).
The CDC model outperforms these baselines because the CDC filter can simultaneously model high-level semantics and temporal action dynamics.
We also evaluate the CDC network with fixed weights in 3D ConvNets and only fine-tune CDC layers, resulting in a minor performance drop. This implies that it is helpful to train CDC networks in an end-to-end manner so that the 3D ConvNets part can be trained to summarize more discriminative information for CDC layers to infer more accurate temporal dynamics.





\begin{table}[t]
\begin{center}
\begin{tabular}{c|ccccc}
IoU threshold & 0.3 & 0.4 & 0.5 & 0.6 & 0.7 \\ \hline
Karaman \etal \cite{th3} &  0.5 &  0.3 &  0.2 &  0.2 &  0.1   \\
Wang \etal \cite{th2} &  14.6 &  12.1 &  8.5 & 4.7 &  1.5   \\
Heilbron \etal \cite{fast_temporal_activity_cvpr16} &  - &  - &  13.5 &  - &  -   \\
Escorcia \etal \cite{victor_eccv16} &  - &  - &  13.9 &  - &  -  \\
Oneata \etal \cite{th1}  &  28.8 &  21.8 &  15.0  &  8.5 &  3.2 \\
Richard and Gall \cite{Richard_2016_CVPR} &  30.0 &  23.2 &  15.2  &  - &  -  \\
Yeung \etal \cite{stanford_cvpr16} &  36.0 &  26.4  &  17.1 &  - &  -  \\
Yuan \etal \cite{yuan_cvpr16} &  33.6 &  26.1 &  18.8  &  - & -   \\
S-CNN  \cite{scnn_shou_wang_chang_cvpr16} &  36.3 &  28.7 & 19.0 &  10.3 &  5.3 \\ \hline
C3D + LinearInterp &  36.0 &  26.4  &  19.6 &  11.1 &  6.6  \\
Conv \& De-conv &  38.6 &  28.2  &  22.4 &  12.0 &  7.5 \\
CDC (fix 3D ConvNets) &  36.9 &  26.2  &  20.4 &  11.3 &  6.8  \\  
\hline
\textbf{CDC} &  \textbf{40.1} & \textbf{ 29.4 }& \textbf{23.3} &  \textbf{13.1} &  \textbf{7.9}
\end{tabular}
\end{center}
\caption{Temporal action localization mAP on THUMOS'14 as the overlap IoU threshold used in evaluation varies from 0.3 to 0.7. - indicates that results are unavailable in the corresponding papers.}
\label{map2}
\end{table}

\subsection{Temporal action localization} \label{loc}

Given per-frame labeling results from the CDC network, we generate proposals, determine class category, and predict precise boundaries following Section \ref{post}.
Our approach is applicable to any segment proposal method.
Here we conduct experiments on THUMOS'14, and thus employ the publicly available proposals generated by the S-CNN proposal network \cite{scnn_shou_wang_chang_cvpr16}, which achieves high recall on THUMOS'14 .
Finally, we follow \cite{yeung2015every,scnn_shou_wang_chang_cvpr16} to perform standard post-processing steps such as non-maximum suppression.


\hfill\break\noindent\textbf{Evaluation metrics.} \label{eval2} Localization performance is also evaluated by mAP. Each item in the rank list is a predicted segment. The prediction is correct when it has the correct category and its temporal overlap IoU with the ground truth is larger than the threshold. Redundant detections for the same ground truth instance are not allowed.

\begin{figure*}[t]
\centering
\includegraphics[width=0.96\textwidth]{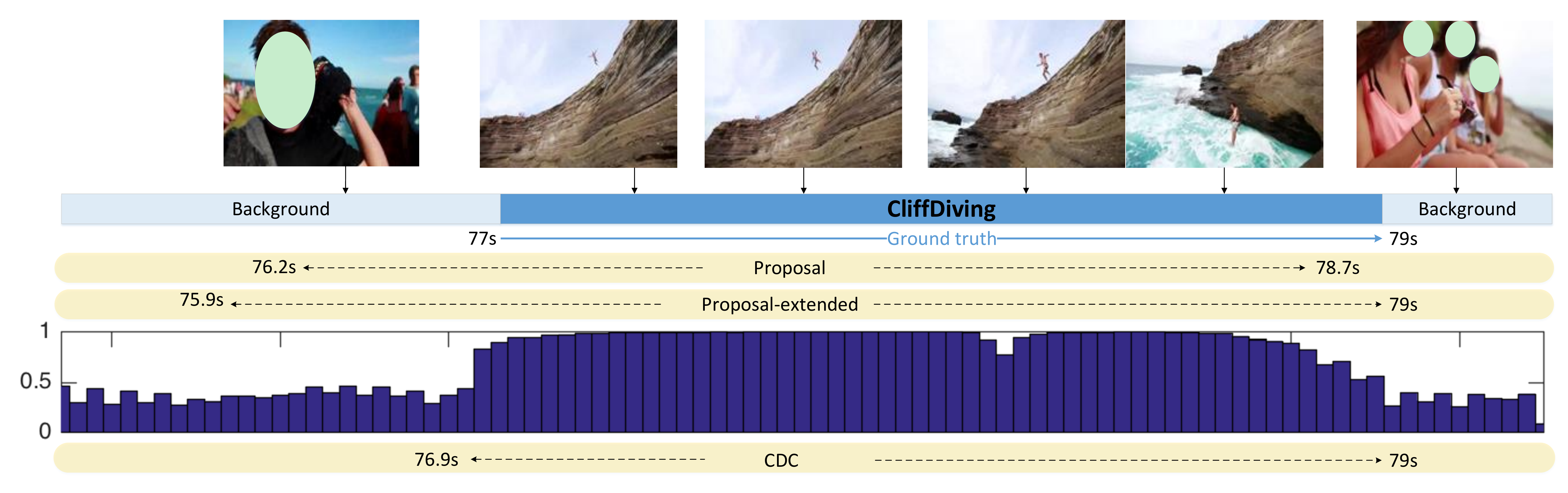}
\caption{Visualization of the process of refining temporal boundaries for a proposal segment. Horizontal axis stands for time. From the top to the bottom: (1) frame-level ground truths for a CliffDiving instance in an input video with some representative frames; (2) a corresponding proposal segment; (3) the proposal segment after extension; (4) the per-frame score of detecting CliffDiving predicted by the CDC network; (5) the predicted action instance after the refinement using CDC.}
\label{refinement}
\end{figure*}

\hfill\break\noindent\textbf{Comparisons.} As shown in Table \ref{map2}, \textbf{CDC} achieves much better results than all the other state-of-the-art methods, which have been reviewed in Section \ref{soa}.
Compared to the proposed CDC model: the typical approach of extracting a set of features to train SVM classifiers and then applying the trained classifiers on sliding windows or segment proposals (Karaman \etal \cite{th3}, Wang \etal \cite{th2}, Oneata \etal \cite{th1}, Escorcia \etal \cite{victor_eccv16}) does not directly address the temporal localization problem.
Systems encoding iDTF with FV (Heilbron \etal \cite{fast_temporal_activity_cvpr16}, Richard and Gall \cite{Richard_2016_CVPR}) cannot learn spatio-temporal patterns directly from raw videos to make predictions.
RNN/LSTM based methods (Yeung \etal \cite{stanford_cvpr16}, Yuan \etal \cite{yuan_cvpr16}) are unable to explicitly capture motion information beyond temporal dependencies.
S-CNN can effectively capture spatio-temporal patterns from raw videos but lacks the ability of adjusting boundaries from proposal candidates. With the proposed CDC filter, the CDC network can determine confidence scores at a fine granularity, beyond segment-level prediction, and hence precisely localize temporal boundaries.
In addition, we employ per-frame predictions of other methods indicated in Table \ref{map1} (C3D + LinearInterp, Conv \& De-conv, CDC with fixed 3D ConvNets
) to perform temporal localization based on S-CNN proposal segments. As shown in Table \ref{map2}, the performance of the CDC network is still better, because more accurate predictions at the same temporal granularity can be used to predict more accurate label and more precise boundaries for the same input proposal segment.
In Figure \ref{refinement}, we illustrate how our model refines boundaries from segment proposal to precisely localize action instance in time.

\subsection{Discussions}

\begin{figure}[h]
\centering
\includegraphics[width=0.43\textwidth]{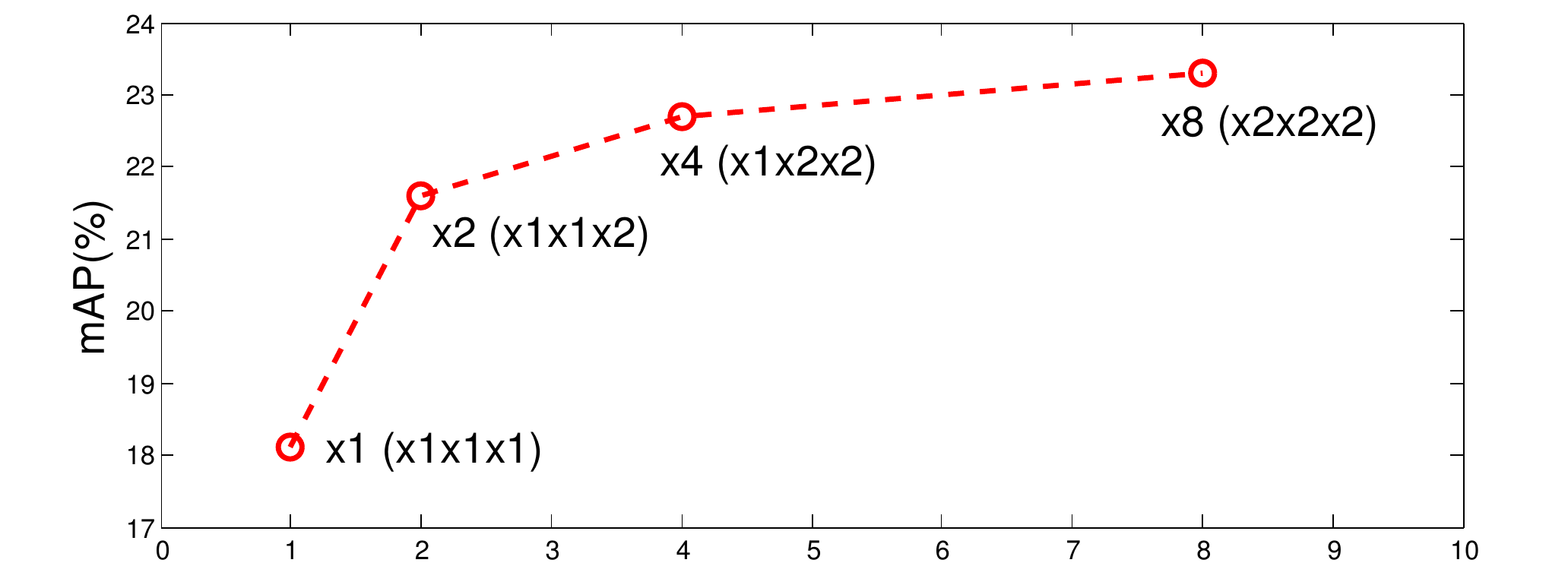}
\caption{mAP gradually increases when the temporal granularity of CDC network prediction increases from x1 (one label for every 8 frames) to x8 (one label per frame).
Each point corresponds to \textbf{x total upscaling factor (x $\tt CDC6$ upscaling factor x $\tt CDC7$ upscaling factor x $\tt CDC8$ upscaling factor)} in time.
We conduct the evaluation on THUMOS'14  with IoU 0.5.}
\label{granularity}
\end{figure}

\noindent\textbf{The necessity of predicting at a fine granularity in time.} In Figure \ref{granularity}, we compare CDC networks predicting action scores at different temporal granularities.
When the temporal granularity increases, mAP increases accordingly. This demonstrates the importance of predicting at a fine-granularity for achieving precise localization.

\hfill\break\noindent\textbf{Efficiency analysis.} The CDC network is compact and demands little storage, because it can be trained from raw videos directly to make fine-grained predictions in an end-to-end manner without the need to cache intermediate features.
A typical CDC network such as the example in Figure \ref{network} only requires around 1GB storage.

Our approach is also fast. Compared with segment-level prediction methods such as S-CNN localization network \cite{scnn_shou_wang_chang_cvpr16}, CDC has to perform more operations due to the need of making predictions at every frame. Therefore, when the proposal segment is long, CDC is less efficient for the sake of achieving more accurate boundaries. But in the case of short proposal segments, since these proposals usually are densely overlapped, segment-level methods have to process a large number of segments one by one.
However, CDC networks only need to process each frame once, and thus it can avoid redundant computations. On a NVIDIA  Titan X GPU of 12GB memory, the speed of a CDC network is around 500 Frames Per Second (FPS), which means it can process a 20s long video clip of 25 FPS within one second.



\begin{table}[b]
\begin{center}
\begin{tabular}{c|ccc|c}
mAP    & 0.5  & \multicolumn{1}{l}{0.75} & \multicolumn{1}{l|}{0.95} & \multicolumn{1}{l}{Average-mAP} \\ \hline
before & 45.1 & 4.1                      & 0.0                       & 16.4                            \\
after  & 45.3 & 26.0                     & 0.2                       & 23.8 
\end{tabular}
\end{center}
\caption{Temporal localization mAP on ActivityNet Challenge 2016 \cite{activitynet} of Wang and Tao \cite{AN1} before and after the refinement step using our CDC network. We follow the official metrics used in \cite{activitynet} to evaluate the average mAP.}
\label{map3}
\end{table}

\hfill\break\noindent\textbf{Temporal activity localization.} Furthermore, we found that our approach is also useful for localizing activities of high-level semantics and complex components.
We conduct experiments on ActivityNet Challenge 2016 dataset \cite{caba2015activitynet,activitynet}, which involves 200 activities, and contains around 10K training videos (15K instances) and 5K validation videos (7.6K instances). Each video has an average of 1.65 instances with temporal annotations. We train on the training videos and test on the validation videos.
Since no activity proposal results of high quality exist, we apply the trained CDC network to the results of the first place winner \cite{AN1} in this Challenge to localize more precise boundaries.
As shown in Table \ref{map3}, they achieve high mAP when the IoU in evaluation is set to 0.5, but mAP drops rapidly when the evaluation IoU increases.
After using the per-frame predictions of our CDC network to refine temporal boundaries of their predicted segments, we gain significant improvements particularly when the evaluation IoU is high (\ie 0.75).
This means that after the refinement, these segments have more precise boundaries and have larger overlap with ground truth instances.

\section{Conclusion and future works}

In this paper, we propose a novel CDC filter to simultaneously perform spatial downsampling (for spatio-temporal semantic abstraction) and temporal upsampling (for precise temporal localization), and design a CDC network to predict actions at frame-level.
Our model significantly outperforms all other methods both in the per-frame labeling task and the temporal action localization task.
Supplementary descriptions of the implementation details and additional experimental results are available in \cite{cdc_zheng_arxiv}.

\section{Acknowledgment}

The project was supported by Mitsubishi Electric, and also by Award No. 2015-R2-CX-K025, awarded by the National Institute of Justice, Office of Justice Programs, U.S. Department of Justice. The opinions, findings, and conclusions or recommendations expressed in this publication are those of the author(s) and do not necessarily reflect those of the Department of Justice. The Tesla K40 used for this research was donated by the NVIDIA Corporation. We thank Wei Family Private Foundation for their support for Zheng Shou, and anonymous reviewers for their valuable comments.

\begin{figure*}[t]
\centering
\includegraphics[width=0.85\textwidth]{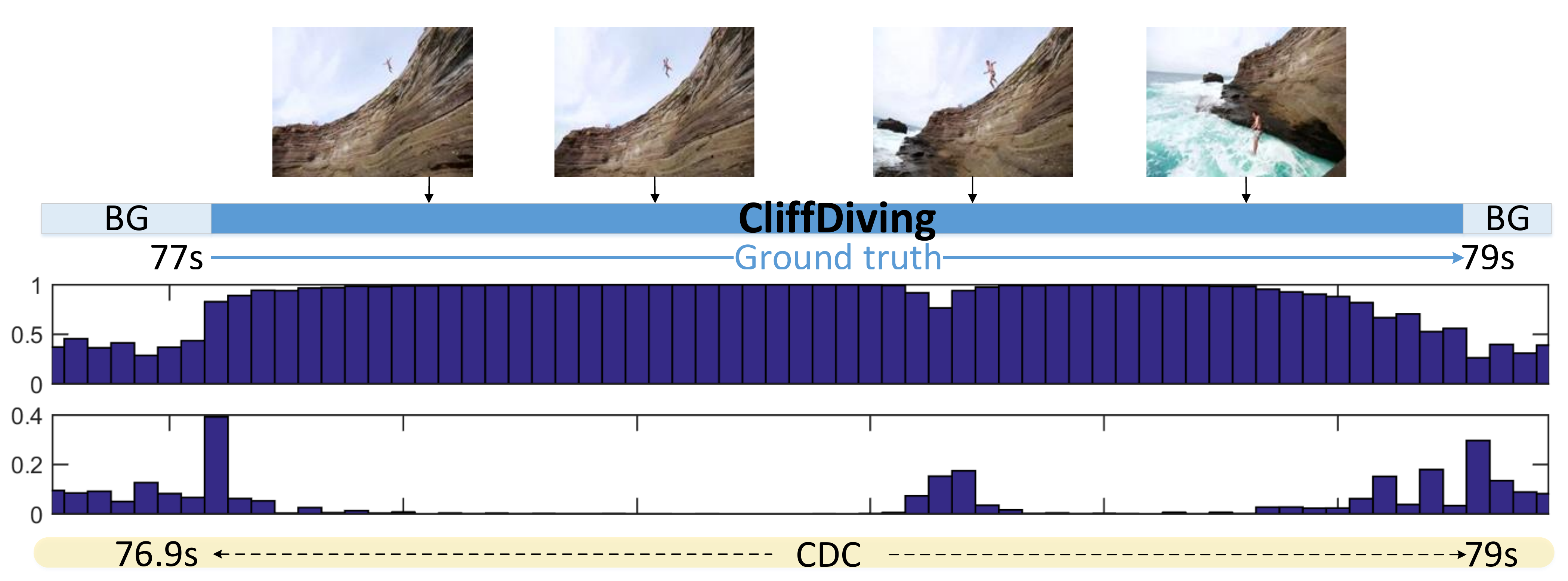}
\caption{We use the frame-level detection scores (3rd row) to compute the absolute frame-to-frame score differences (4th row), which show high correlations with the true action boundaries.}
\label{gradient}
\end{figure*}

\section{Appendix}

\subsection{Additional justification of the motivation}

As mentioned in the paper, the traditional approaches use segment-level detection, in which segment proposals are analyzed to predict the action class in each segment. Such approaches are limited by the fixed segment lengths and boundary locations, and thus inadequate for finding precise action boundaries. Here we proposed a novel model to first predict actions at fine-level and then use such fine-grained score sequences to accurately detect the action boundaries. The fine-grained score sequence also offers natural ways to determine the score threshold needed in refining boundaries at the frame level. Also, though not emphasized in the paper, the fine-level score sequence can also be used to select precise keyframes or discover sub-actions within an action.

Following the reviewer's suggestion, we also computed the frame-to-frame score gradient using the frame-level detection results. As shown in Figure \ref{gradient}, the frame-level gradient peaks nicely correlate with the action boundaries, confirming the intuition of using the fine-level detection results.
Also, as shown in Figure \ref{granularity} in the paper, when the temporal granularity increases, localization performance increases accordingly.
Finally, our motivation is quantitatively justified by the good results on two standard benchmarks as shown in Section \ref{exp}.

\subsection{Additional implementation details}

\noindent\textbf{Temporal boundary refinement.} Here, we provide details and pseudo-codes for temporal boundary refinement presented in Section \ref{post}. Algorithm \ref{alg-refine} is used to refine boundaries of each proposal segment.
Also, our source codes can be found at \url{https://bitbucket.org/columbiadvmm/cdc}.


\begin{algorithm}[H]
\textbf{Input}: A proposal segment of starting frame index $t_s$ and ending frame index $t_e$, the percentage parameter of segment length expansion  $\alpha$, the first frame index $v_s$ and the last frame index $v_e$ of the video containing the proposal segment, the total number of categories $K$

\textbf{Output}: the refined starting frame index ${t_s}'$ and ending frame index ${t_e}'$, the predicted category $c$, the predicted confidence score $s$

	1. // Extend boundaries on both sides by the percentage of the original segment length
	
	2. ${t_s}' = \max \left( {{v_s},{t_s} - \alpha  \cdot \left( {{t_e} - {t_s} + 1} \right)} \right)$
	
	3. ${t_e}'= \min \left( {{v_e},{t_e} + \alpha  \cdot \left( {{t_e} - {t_s} + 1} \right)} \right)$
	
	4. // Feed frames into the CDC network to produce the confidence score matrix ${\bf{P}} \in {\Re^{\left( {{t_e} - {t_s} + 1} \right) \times K}}$
	
	5. $\bf{P} =$ \textbf{CDC}(frames from ${t_s}'$ to ${t_e}'$)
	
	6. assign $c$ as the category with the maximum average confidence score over all frames from ${t_s}'$ to ${t_e}'$
	
	7. // Estimate the mean $\mu$ and the standard deviation $\sigma$
	
	8. $\mu ,\sigma  =$ Gaussian Kernel Density Estimation$\left( {{\bf{P}}\left[ {:,c} \right]} \right)$ 
	
	9. $\beta  = \mu  - \sigma $ \quad // Compute the score threshold
	
	10. // Refine the starting time
	
	11. \textbf{for} $i_s = 1,2, \ldots , \left( {t_e}'-{t_s}'+1 \right)  $ \textbf{do} 
	
	12. \quad \textbf{if} ${\bf{P}}\left[ {i_s,c} \right] >  = \beta$ \textbf{then} 
	
	13. \quad \quad \textbf{break}
	
	14. \quad \textbf{end if}
	
    15. \textbf{end for}

	16. // Refine the ending time
	
	17. \textbf{for} $i_e =  \left( {t_e}'-{t_s}'+1 \right) ,\ldots , 2,1 $ \textbf{do} 
	
	18. \quad \textbf{if} ${\bf{P}}\left[ {i_e,c} \right] >  = \beta$ \textbf{then} 
	
	19. \quad \quad \textbf{break}
	
	20. \quad \textbf{end if}
	
    21. \textbf{end for}

	22. ${t_e}' = {t_s}' + {i_e} - 1$
	
	23. ${t_s}' = {t_s}' + {i_s} - 1$
    
    24. $s = \frac{{\sum\limits_{i = {i_s}}^{{i_e}} {{\bf{P}}\left[ {i,c} \right]} }}{{{t_e} - {t_s} + 1}}$ \quad // Compute the average score
    
    25. \textbf{return} ${t_s}'$, ${t_e}'$, $c$, $s$

\caption{Temporal Boundary Refinement}
\label{alg-refine}
\end{algorithm}

\hfill\break\noindent\textbf{Discussions about the window length used during creating mini-batches.} During mini-batch construction, ideally we would like to set the window length as longer as possible. Therefore, when CDC processes each window, it can take into account more temporal contextual information. However, due to the limitation of the GPU memory, if the window length is too high, we have to set the number of training samples for each mini-batch to be very small, which will make the optimization unstable and thus the training procedure cannot converge well. Also, a long window usually contains much more background frames than action frames and thus we need to further handle the data imbalance issue. During experiments, we conduct a grid search of window length in 16, 32, 64, 128, 256, 512 and empirically found that setting the window length to 32 frames is a good trade-off on a single NVIDIA  Titan X GPU of 12GB memory: (1) we can include sufficient temporal contextual information to achieve good accuracy and (2) we can set the batch size as 8 to guarantee stable optimization.

\begin{figure*}[t]
\centering
\includegraphics[width=0.85\textwidth]{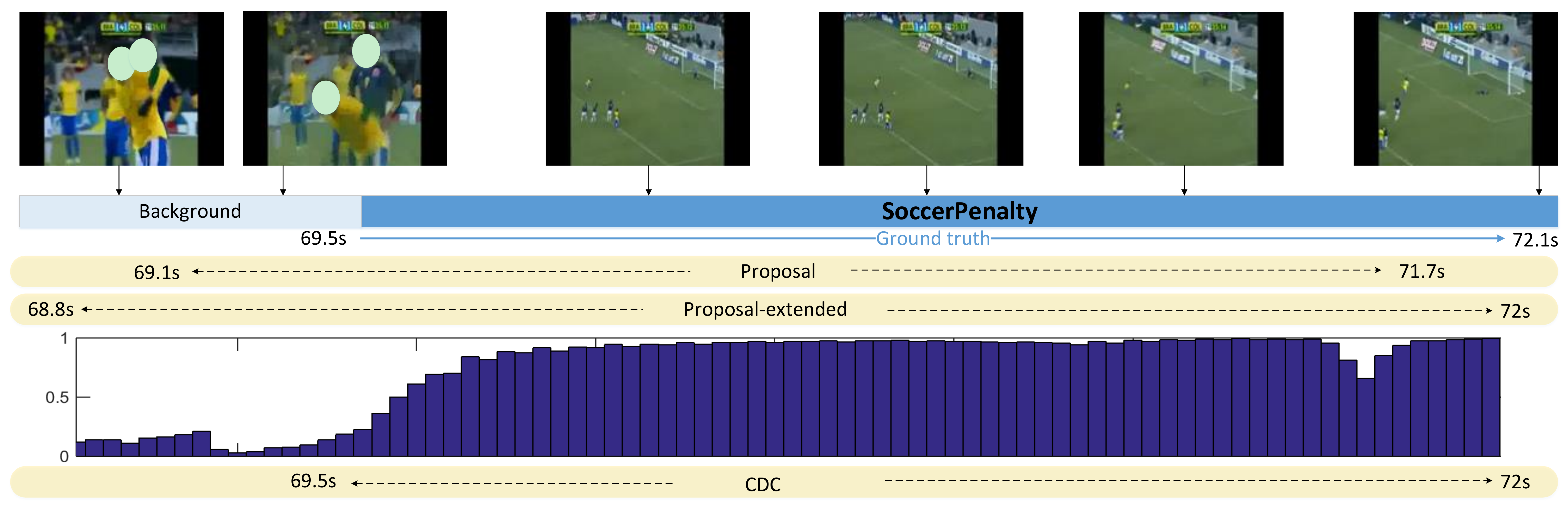}
\caption{Visualization of the process of refining temporal boundaries for an action proposal segment. Horizontal axis stands for time. From the top to the bottom: (1) frame-level ground truths for a SoccerPenalty action instance in a test video with some representative frames; (2) a corresponding proposal segment; (3) the proposal segment after extension; (4) the per-frame score of being SoccerPenalty predicted by the CDC network; (5) the precisely predicted action instance after the refinement step using CDC.}
\label{refinement1}
\end{figure*}

\begin{figure*}[t]
\centering
\includegraphics[width=0.85\textwidth]{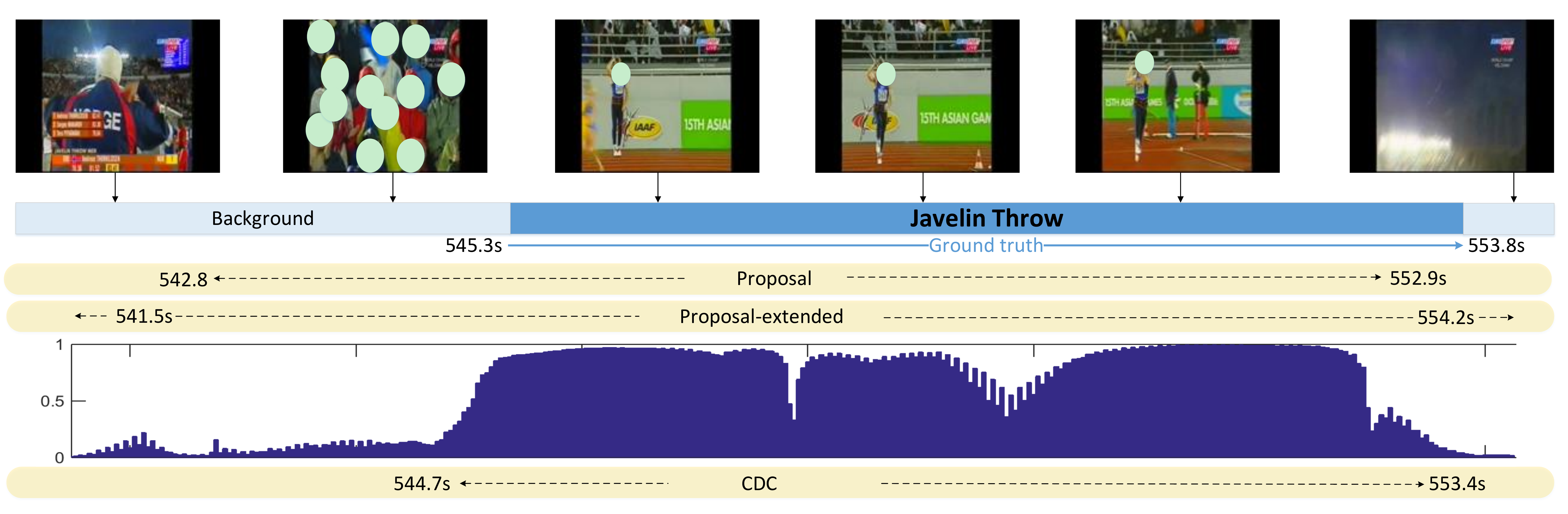}
\caption{Visualization of the process of refining temporal boundaries for an action proposal segment. Horizontal axis stands for time. From the top to the bottom: (1) frame-level ground truths for a JavelinThrow action instance in a test video with some representative frames; (2) a corresponding proposal segment; (3) the proposal segment after extension; (4) the per-frame score of being JavelinThrow predicted by the CDC network; (5) the precisely predicted action instance after the refinement step using CDC.}
\label{refinement2}
\end{figure*}

\subsection{Additional experiments}

\noindent\textbf{Sensitivity analysis.} When we extend the segment proposal, the percentage $\alpha$ of the original proposal length should not be too small so that our model can consider a wider interval and not be too large to include too many irrelevant frames. As shown in Table \ref{map4}, the system has stable performances when $\alpha$ varies within a reasonable range.

\begin{table}[h]
\begin{center}
\begin{tabular}{c|ccccc}
$\alpha$ & 1/8 & 1/7 & 1/6 & 1/5 & 1/4 \\ \hline
mAP & 23.3 &  23.2 &  23.1 & 23.1  &  23.6
\end{tabular}
\end{center}
\caption{mAP on THUMOS'14  with the evaluation IoU set to 0.5 when we vary the extension percentage $\alpha$ of the original proposal length from 1/8 to 1/4.}
\label{map4}
\end{table}

\hfill\break\noindent\textbf{Additional results on ActivityNet.} We expand the comparisons on ActivityNet validation set to include results provided by additional top performers [51, 52] in ActivityNet Challenge 2016. As shown in Table \ref{map3}, our method CDC outperforms all other methods. As shown in Table \ref{map3test}, CDC also performs the best on ActivityNet test set.

\begin{table}[h]
\begin{center}
\begin{tabular}{c|ccc|c}
IoU threshold    & 0.5  & \multicolumn{1}{l}{0.75} & \multicolumn{1}{l|}{0.95} & \multicolumn{1}{l}{Ave-mAP} \\ \hline
Singh and Cuzzolin \cite{AN2} & 22.7 & 10.8                      & 0.3                       & 11.3                            \\
Singh \cite{AN3} & 26.0 & 15.2                      & 2.6                       & 14.6                            \\
Wang and Tao \cite{AN1} & 45.1 & 4.1                      & 0.0                       & 16.4                            \\ \hline
CDC  & 45.3 & 26.0                     & 0.2                       & 23.8 
\end{tabular}
\end{center}
\caption{Additional baseline results of temporal localization mAP on ActivityNet Challenge 2016 \cite{activitynet} validation set. The baseline results are kindly provided by the authors of \cite{AN2,AN3,AN1}.}
\label{map3}
\end{table}

\begin{table}[h]
\begin{center}
\begin{tabular}{c|ccc|c}
IoU threshold    & 0.5  & \multicolumn{1}{l}{0.75} & \multicolumn{1}{l|}{0.95} & \multicolumn{1}{l}{Ave-mAP} \\ \hline
Singh and Cuzzolin \cite{AN2} & 36.4 & 11.1                      & 0.1                       & 17.8                            \\
Singh \cite{AN3} & 28.7 & 17.8                      & 2.9                       & 17.7                            \\
Wang and Tao \cite{AN1} & 42.5 & 2.9                      & 0.1                       & 14.6                            \\ \hline
CDC  (train) & 43.1 & 25.6                     & 0.2                       & 22.9 \\
CDC  (train+val) & 43.0 & 25.7                     & 0.2                       & 22.9 
\end{tabular}
\end{center}
\caption{Comparisons of temporal localization mAP on ActivityNet Challenge 2016 \cite{activitynet} test set. The baseline results are quoted from the ActivityNet Challenge 2016 leaderboard \cite{activitynet}. CDC  (train) is training the CDC model on the training set only and CDC  (train+val) uses the training set and the validation set together to train the CDC model. }
\label{map3test}
\end{table}

\hfill\break\noindent\textbf{Discussions about other proposal methods.} As shown in Table \ref{map2s}, we evaluate temporal localization performances of CDC based on other proposals on THUMOS'14.



\begin{table*}[t]
\begin{center}
\begin{tabular}{c|ccccc}
IoU threshold & 0.3 & 0.4 & 0.5 & 0.6 & 0.7 \\ \hline
S-CNN  \cite{scnn_shou_wang_chang_cvpr16} w/o CDC &  36.3 &  28.7 & 19.0 &  10.3 &  5.3 \\
ResC3D+S-CNN \cite{scnn_shou_wang_chang_cvpr16}  w/o CDC &   40.6 & 32.6 & 22.5 &  12.3 &  6.4 \\ \hline
S-CNN \cite{scnn_shou_wang_chang_cvpr16} &   40.1 & 29.4 & 23.3 &  13.1 &  7.9 \\
ResC3D+S-CNN  \cite{scnn_shou_wang_chang_cvpr16} &  41.3 &  30.7 & 24.7 &  14.3 &  8.8
\end{tabular}
\end{center}
\caption{Temporal action localization mAP on THUMOS'14 as the overlap IoU threshold used in evaluation varies from 0.3 to 0.7. We evaluate our CDC model based on different proposal methods.}
\label{map2s}
\end{table*}

On ActivityNet, the proposals currently used in Section \ref{exp} from \cite{AN1} is a reasonable choice - its recall is 0.681 with 56K proposals when evaluate at IoU=0.5 on the validation set. We also have considered using other state-of-the-art proposal methods: (1) The ActivityNet challenge provides proposals computed by \cite{fast_temporal_activity_cvpr16}, but it has a low recall at 0.527 on the validation set with 441K proposals, which contain a lot of false alarms. (2) DAPs \cite{victor_eccv16} advocates that train proposal model on THUMOS and then generalize the model to ActivityNet. Due to lack training data from ActivityNet, DAPs has a quite low recall at around 0.23 and is not a reasonable proposal candidate. (3) S-CNN \cite{scnn_shou_wang_chang_cvpr16} is designed for instance-level detection. However, ground truth annotations in ActivityNet do not distinguish consecutive instances - one ground truth interval can contain multiple activity instances. Also, for activities of high-level semantics, it is ambiguous to define what is an individual activity instance. Therefore, S-CNN does not suit ActivityNet.

\hfill\break\noindent\textbf{Additional discussions about speed.} For the sake of avoiding confusions, we would like to emphasize that the CDC network is end-to-end while the task of temporal localization is not end-to-end due to the need of combing with proposals and performing post-processing. Throughout the paper, the speed is also computed for the CDC network itself.
Following C3D \cite{3dcnn}, each input frame has spatial resolution $128 \times 171$ and will be cropped into $112 \times 112$ as network input (random cropping during training and center cropping during testing). As indicated in Figure \ref{network}, each input video of $L$ frames has the shape of (3, $L$, 112, 112).
As aforementioned, on a single NVIDIA  Titan X GPU of 12GB memory, the speed of a CDC network is around 500 Frames Per Second (FPS), which means it can process a 20s long video clip of 25 FPS within one second.

\hfill\break\noindent\textbf{Additional visualization examples.} As supplementary material to Figure \ref{refinement}, we provide additional examples to show the process of using Convolutional-De-Convolutional (CDC) model to refine the boundaries of proposal segments and achieve precise temporal action localization on THUMOS'14 \cite{THUMOS14}.
As shown in Figure \ref{refinement1} and Figure \ref{refinement2}, the combination of the segment proposal and the CDC frame-level score prediction is powerful.
The segment proposal allows for leveraging candidates of coarse boundaries to help handle the noisy outliers in the dipped score intervals such as shown in Figure \ref{refinement2}.
The proposed CDC model allows for fine-grained predictions at the frame level to help refine the segment boundaries in frame-level for precise localization.

{\small
\bibliographystyle{ieee}
\bibliography{egbib}
}

\end{document}